\documentclass{article}

     \PassOptionsToPackage{round,authoryear}{natbib}

 \usepackage[preprint]{style}

\usepackage[utf8]{inputenc} 
\usepackage[T1]{fontenc}    
\usepackage{hyperref}       
\usepackage{url}            
\usepackage{booktabs}       
\usepackage{amsfonts}       
\usepackage{nicefrac}       
\usepackage{microtype}      
\usepackage{xcolor}         
\usepackage{bbm}
\usepackage{amsmath}
\usepackage{multirow}
\usepackage[normalem]{ulem}
\usepackage{ulem}
\usepackage{graphics}
\useunder{\uline}{\ul}{}
\usepackage{graphicx}
\usepackage{wrapfig}
\usepackage{amssymb}
\usepackage{tabularx}
\usepackage{fancyvrb}
\usepackage{listings}
\usepackage{authblk}

\usepackage{scrextend}
\usepackage{xspace}
\usepackage{longtable}

\usepackage{calligra}
\usepackage[T1]{fontenc}
\usepackage{eucal}
\usepackage{mathrsfs}
\usepackage{fontawesome5}

\def\modelname{\texttt{Voila}\xspace}
\def\modelnameete{\texttt{Voila-e2e}\xspace}
\def\modelnameauto{\texttt{Voila-autonomous}\xspace}
\def\modelnametoken{\texttt{Voila-Tokenizer}\xspace}

\RequirePackage{fancyhdr}
\title{\Large Voila: Voice-Language Foundation Models for\\
Real-Time Autonomous Interaction and Voice Role-Play}

\author{Yemin Shi\textsuperscript{*}}
\author{Yu Shu\textsuperscript{*}}
\author{Siwei Dong\textsuperscript{*}}
\author{Guangyi Liu\textsuperscript{*}}
\author{Jaward Sesay}
\author{Jingwen Li}
\author{Zhiting Hu}

\affil{Maitrix.org,~~ UC San Diego,~~ MBZUAI \\
    \textsuperscript{*}Equal contribution}

\begin{document}
\maketitle

\begin{abstract}

A voice AI agent that blends seamlessly into daily life would interact with humans in an autonomous, real-time, and emotionally expressive manner. Rather than merely reacting to commands, it would continuously listen, reason, and respond proactively, fostering fluid, dynamic, and emotionally resonant interactions. We introduce \modelname, a family of large voice-language foundation models that make a step towards this vision. 
Voila moves beyond traditional pipeline systems by adopting a new end-to-end architecture that enables full-duplex, low-latency conversations while preserving rich vocal nuances such as tone, rhythm, and emotion. It achieves a response latency of just 195 milliseconds, surpassing the average human response time. Its hierarchical multi-scale Transformer integrates the reasoning capabilities of large language models (LLMs) with powerful acoustic modeling, enabling natural, persona-aware voice generation---where users can simply write text instructions to define the speaker’s identity, tone, and other characteristics. Moreover, \modelname supports over one million pre-built voices and efficient customization of new ones from brief audio samples as short as 10 seconds. Beyond spoken dialogue, \modelname is designed as a unified model for a wide range of voice-based applications, including automatic speech recognition (ASR), Text-to-Speech (TTS), and, with minimal adaptation, multilingual speech translation. \modelname is fully open-sourced to support open research and accelerate progress toward next-generation human-machine interactions.
\end{abstract}

\begin{table}[!h]
    \centering
    \begin{tabular}{l l}
        \includegraphics[width=0.028\textwidth]{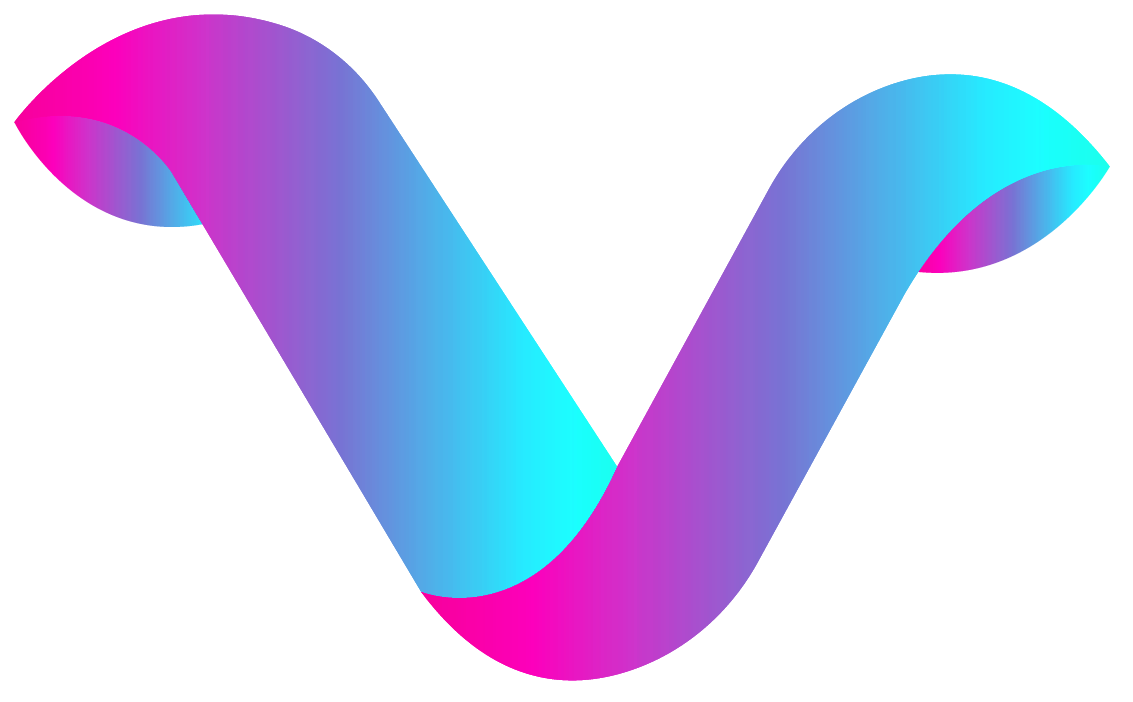} \,Voila Project Page & \href{https://voila.maitrix.org}{\color{blue} voila.maitrix.org} \\
        \,{\faGlobe} \,Voila Demo & \href{https://hf.co/spaces/maitrix-org/Voila-demo}{\color{blue} hf.co/spaces/maitrix-org/Voila-demo} \\
        \includegraphics[width=0.028\textwidth]{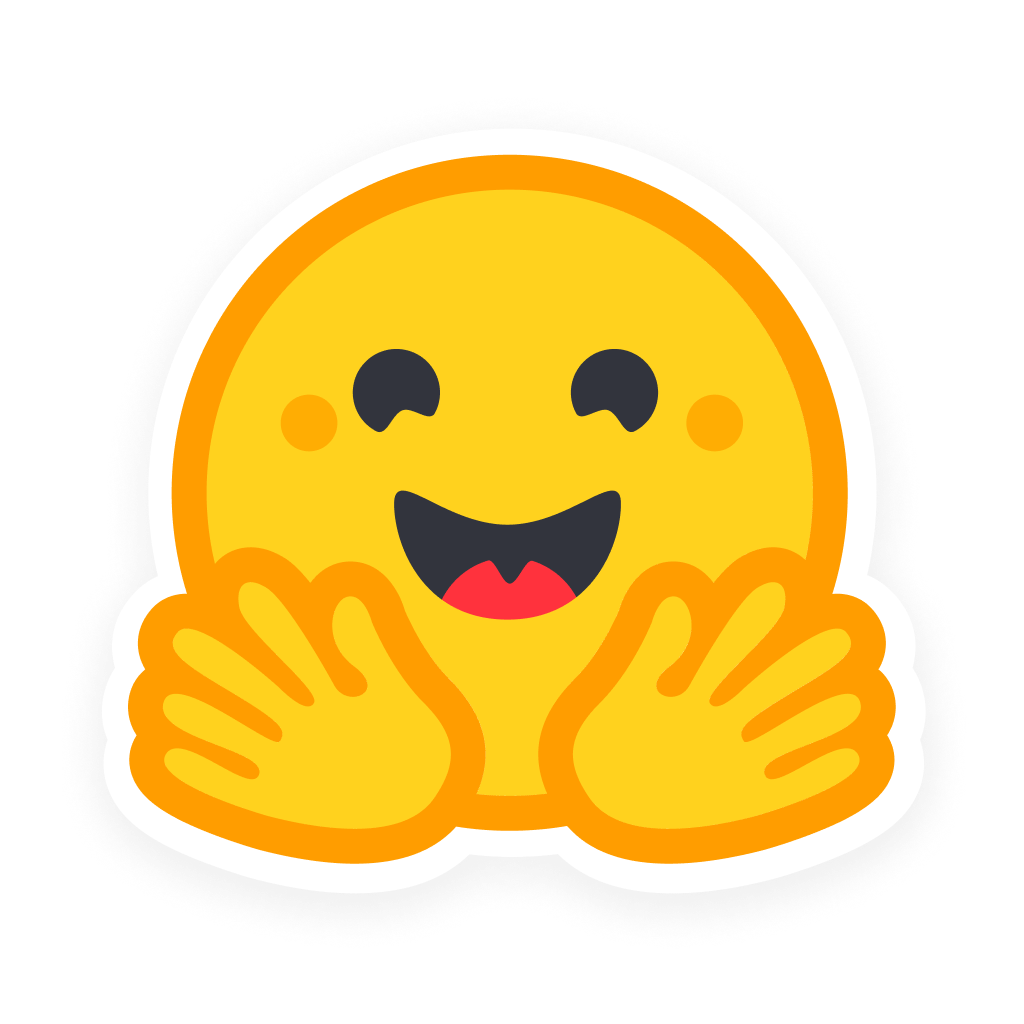} Voila Base Model & \href{https://hf.co/maitrix-org/Voila-base}{\color{blue} hf.co/maitrix-org/Voila-base} \\
        \includegraphics[width=0.028\textwidth]{Figures/hf-logo.png} Voila End-to-end Model & \href{https://hf.co/maitrix-org/Voila-chat}{\color{blue} hf.co/maitrix-org/Voila-chat} \\
        \includegraphics[width=0.028\textwidth]{Figures/hf-logo.png} Voila Full-duplex Model (preview) & \href{https://hf.co/maitrix-org/Voila-autonomous-preview}{\color{blue} hf.co/maitrix-org/Voila-autonomous-preview} \\
        \includegraphics[width=0.028\textwidth]{Figures/hf-logo.png} Voila Tokenizer & \href{https://hf.co/maitrix-org/Voila-Tokenizer}{\color{blue} hf.co/maitrix-org/Voila-Tokenizer} \\
        \includegraphics[width=0.028\textwidth]{Figures/hf-logo.png} Voila Benchmark & \href{https://hf.co/datasets/maitrix-org/Voila-Benchmark}{\color{blue} hf.co/datasets/maitrix-org/Voila-Benchmark} \\
        \includegraphics[width=0.028\textwidth]{Figures/hf-logo.png} Voila Voice Library & \href{https://hf.co/datasets/maitrix-org/Voila-million-voice}{\color{blue} hf.co/datasets/maitrix-org/Voila-million-voice} \\
        \,{\faGithub} \,Voila Code & \href{https://github.com/maitrix-org/Voila}{\color{blue} github.com/maitrix-org/Voila}
    \end{tabular}
\end{table}

\section{Introduction}\label{sec:intro}

\paragraph{Autonomous Interaction} 
Most AI systems today interact with humans reactively: they wait passively and respond after receiving a user query. For instance, in conversational systems ranging from Siri to ChatGPT, the user asks a question, the system generates an answer, and then waits for the next prompt, resulting in rigid, turn-based interactions. While this command-driven scheme may be sufficient for basic AI assistants, it remains far from achieving a truly {\it autonomous} machine capable of engaging proactively and emulating the rich dynamics of human-to-human interaction. An autonomous AI would {\it continuously assess its context, anticipate user needs in real-time, and determine if, when, and how to interact in an optimal way} \citep{buss2012autonomous,grosinger2022proactive,buyukgoz2022two,hoke2021digital}. For example, when a user is walking down a street, the AI might warn them about an approaching cyclist they had not noticed, or suggest a stop at a hidden gem caf\'e nearby. Similarly, if a user keeps expressing a low mood and spiraling into negative thoughts, the system might actively interrupt to suggest a calming activity tailored to the user’s emotional needs, instead of passively waiting for the user to ask for help. This vision of autonomous interaction has been imagined in popular culture, as seen in the film {\it Her}, where AI interacts fluidly with humans, forming genuine and emotive bonds. Such autonomous interactions make AI more than just a passive tool, but a trusted companion and teammate that blends seamlessly into our daily life.

Among the various modes of communication---text, vision, and gestures---{\it voice} is perhaps the most essential and natural for carrying out autonomous interactions \citep{schafer1995scientific,flanagan1972speech}. Unlike text-based communication, which is often static, asynchronous, and turn-based, voice naturally enables rich, dynamic, and human-like interactions. 
For example, we speak to draw attention and initiate dialogue (even when the other person is not looking), interrupt or overlap speech to signal urgency or redirect conversational flow, and use simple backchanneling cues like `\texttt{mmh}' or `\texttt{yeah}' to convey attentiveness and engagement when others speak \citep{skantze2021turn,yang2022multimodal}. In addition, voice carries rich vocal cues (such as tone, inflection, and rhythm) and subtle emotional nuances that other modalities cannot replicate \citep{bora2024breaking,schroeder2016mistaking}. As a result, a voice-based interface is crucial for creating an engaging and immersive experience in human-machine interactions.

\begin{figure}[t]
    \centering
    \includegraphics[width=0.9\textwidth]{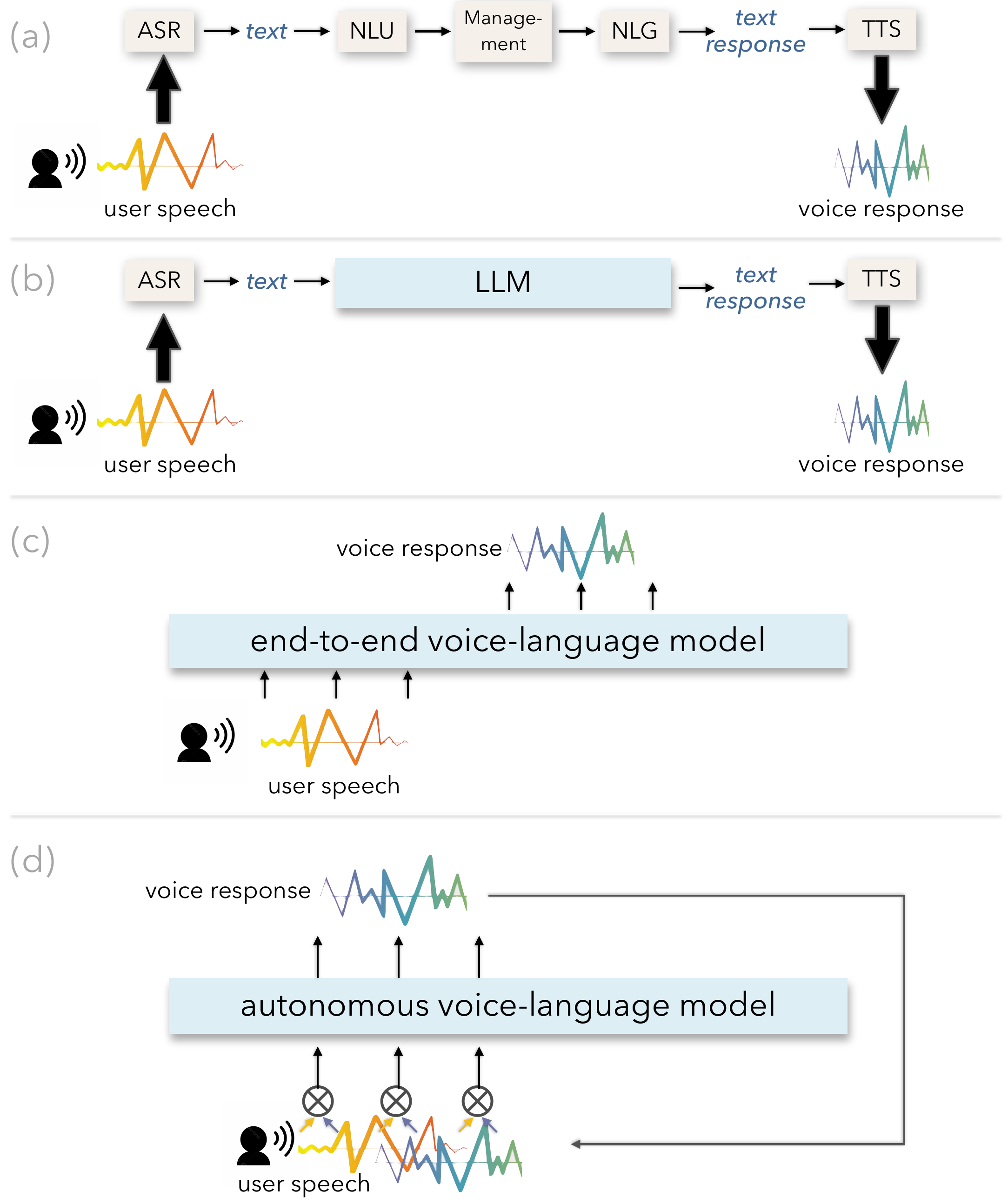}
    \caption{Different paradigms of voice conversation systems: {\bf (a)} Traditional pipeline systems, such as Apple Siri, Amazon Alexa, and Google Assistant, launched in the 2010s; {\bf (b)} Simplified pipeline systems using LLMs to handle text-based understanding and response generation; {\bf (c)} End-to-end audio-in, audio-out systems that offer low latency and rich vocal nuances; {\bf (d)} Autonomous systems that further enable dynamic, proactive interactions.}
    \label{fig:voice-ai}
\end{figure}

\paragraph{Voice AI}
From Bell Labs' Audrey in 1952, which could recognize the sound of a spoken digit from 0 to 9, to recent advancements like ChatGPT-4o \citep{hurst2024gpt} for open-ended spoken dialogue, voice AI has undergone a remarkable evolution, as illustrated in Figure \ref{fig:voice-ai}. 
Early voice systems like Apple Siri, Amazon Alexa, and Google Assistant, launched in the 2010s, pioneered the first widely used voice conversational interfaces and relied on complex modular pipelines (Figure \ref{fig:voice-ai}a). The components in these systems require extensive hand-engineering and are limited to processing a constrained set of user queries. Recent large language models (LLMs) have ushered in a simpler pipeline design that can support open-ended conversations (Figure \ref{fig:voice-ai}b). This new approach consists of three main components: automatic speech recognition (ASR) for converting human speech into text, LLM for generating text responses, and text-to-speech (TTS) for converting the text responses to audio. 
The core of this pipeline is the text-based conversation managed by the LLM, while ASR and TTS work to convert voice into text and text back into voice, enabling audio interfaces.

This pipeline approach leverages the strengths of LLMs in text-based interaction, such as broad knowledge, complex reasoning \citep{wei2022chain,yao2022react,hao2023reasoning,hao2023toolkengpt}, instruction following \citep{ouyang2022training}, and role playing \citep{wang2023rolellm,shao2023character,wu2024role}. However, the pipeline design also comes with fundamental limitations that hinder truly natural, human-like voice interactions, including {\it (1) high latency, (2) loss of vocal nuances, and (3) reactive, turn-based interactions}. Specifically, each module in the pipeline can introduce delays, which often accumulate to several seconds---significantly longer than the average human response time of under 300 milliseconds \citep{stivers2009universals,meyer2023timing}. Additionally, converting audio to text for LLM processing results in a loss of rich acoustic cues, such as tone, accent, emotion, and background sounds. Without these nuances, the system struggles to interpret and respond naturally to the subtleties of human communication---for example, the phrase ``\texttt{Oh, really}'' might express surprise or disinterest depending on the speaker’s tone. 
Moreover, the pipeline inherits the rigid, turn-based structure of text-based conversation mentioned earlier, where speakers respond only after the other has finished. The approach thus fails to capture the essence of natural, autonomous voice interactions. For instance, common conversational elements such as backchanneling, interruptions, and overlapping speech are absent, resulting in interactions that feel mechanical rather than unfolding organically with spontaneity, dynamism, and engagement.
While additional components can be added in the pipeline, such as interruption detection to let users interrupt the system's responses \citep{lebourdais2024automatic} or various triggers that activate the system to initiate interactions under specific conditions (e.g., at a user-specified time) \citep{berube2024proactive}, such rule-based controls are inherently limited. They lack deeper contextual understanding and autonomy necessary to achieve natural, dynamic interactions.

Recent advancements have further led to {\it end-to-end} audio-language models that bypass the traditional pipeline architecture (Figure \ref{fig:voice-ai}c). Instead of converting audio signals into text, this approach processes audio representations (e.g., audio tokens) directly within a large model, which then generates a response in the same audio representation space before decoding it into a voice output. By removing text as an intermediate step, the method preserves rich acoustic details in the input and enables nuanced acoustic generation in the output, permitting more natural and engaging interactions. In addition, without the cascading delays of a multi-module pipeline, end-to-end models can achieve lower latency. 
On the other hand, however, the system still adheres to a reactive, turn-based interaction flow as in prior approaches. 

\paragraph{Voila for Autonomous Voice Interaction} We introduce \modelname, a family of large audio-language foundation models aiming to overcome the above challenges and enable real-time, natural, and flexible voice interaction. Specifically, \modelnameete is an end-to-end model for natural voice conversations with low latency, rich vocal details, and strong instruction following and customizability. 
\modelnameauto further aims for autonomous interaction, where the model continuously listens, reasons, and responds in a full-duplex and simultaneous manner (Figure \ref{fig:voice-ai}d), delivering a next-level voice interaction experience. 

\begin{figure}[t]
    \centering
    \includegraphics[width=0.9\textwidth]{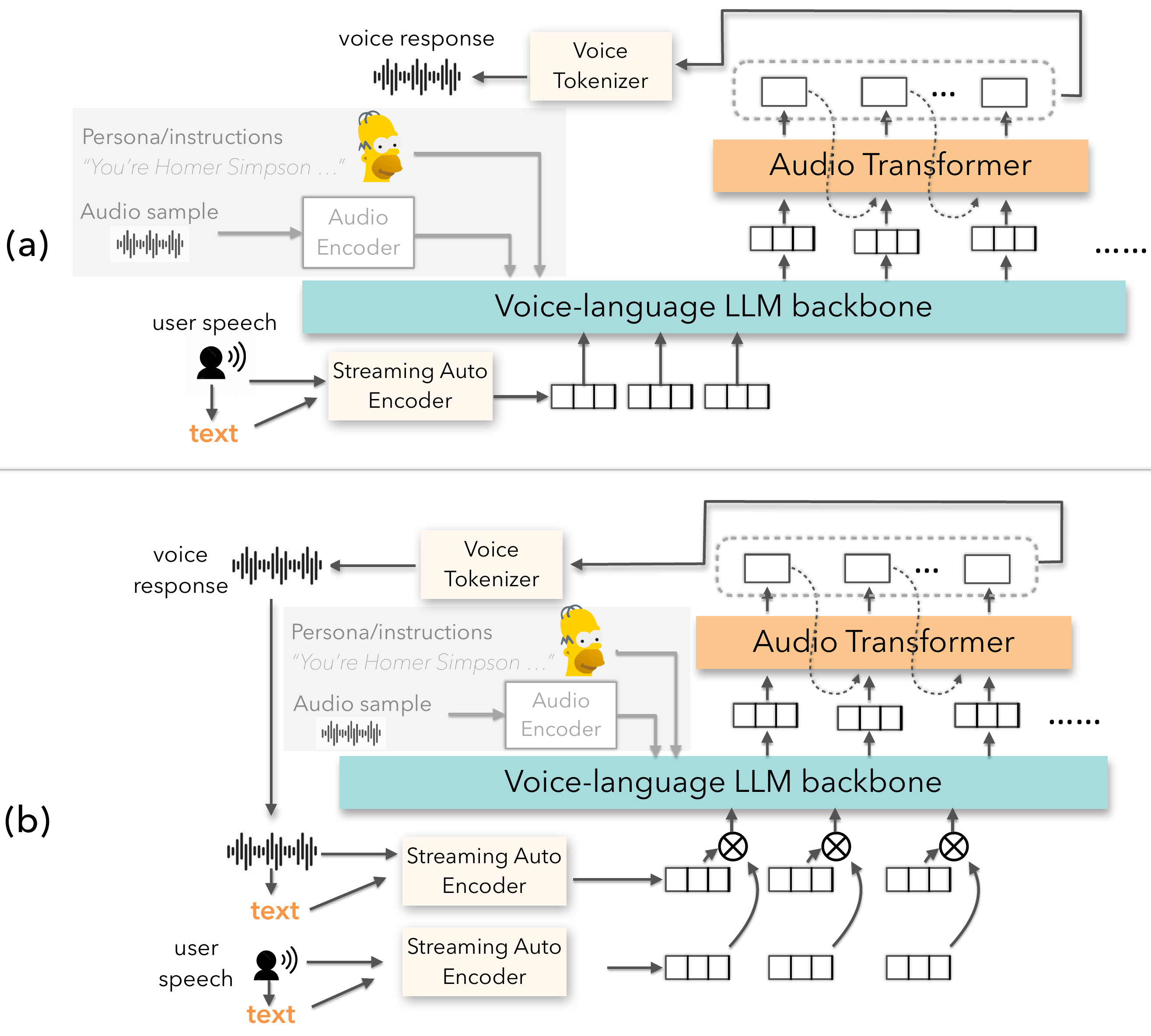}
    \caption{\modelname models: {\bf (a)} \modelnameete for end-to-end voice conversation, {\bf (b)} \modelnameauto for autonomous interaction. Both models allow easy customization of speaker characteristics and voice via text instructions and audio samples.}
    \label{fig:viola-arch-joint}
\end{figure}

\modelname designs a hierarchical Transformer architecture, including streaming audio encoding and tokenization, and multi-scale Transformers consisting of an LLM backbone and a hierarchical audio generator (Figure~\ref{fig:viola-arch-joint}). The models are trained in an end-to-end way with extensive audio-text data.
In summary, \modelname features the following key advancements:
\begin{addmargin}[-26pt]{0pt} 
\begin{itemize}\setlength\itemsep{0pt}

\item {\bf Effective integration of voice and language modeling capabilities}: 
\modelname adopts a range of designs to best combine the text-based capabilities of the pre-trained LLM and the newly learned voice modeling capabilities. For instance, much like how users can input a text prompt to instruct an LLM's behavior, \modelname allows users to do the same on its backbone LLM to define its persona and steer its responses in voice conversations. In addition, \modelname's voice mode retains the extensive knowledge and linguistic proficiency acquired during LLM pretraining, ensuring high-quality model responses. 

To this end, \modelname uses a multi-scale Transformer architecture \citep{yang2023uniaudioaudiofoundationmodel,zhu2024generativepretrainedspeechlanguage,defossez2024moshi} that predicts semantic and acoustic tokens separately at different levels. The disentanglement allows the backbone LLM to focus on handling the semantic information as it was pretrained for, while delegating acoustic information modeling to other Transformer modules. To extract semantic and acoustic tokens from audio data, we build \modelnametoken, a neural audio codec \citep{zeghidour2021soundstream,kumar2024high,zhang2023speechtokenizer}. These audio tokens are added to text vocabulary for cross-modal training and knowledge sharing between modalities. In addition, \modelname interleaves audio and text tokens during generation, leveraging the backbone LLM's text generation capabilities to guide the production of coherent voice responses.

\item {\bf Millions of pre-built voices and efficient voice creation}:  
\modelname allows users to easily customize and plug in new voices for conversations. Given an audio clip of any length (from a few seconds to several hours), \modelname learns a voice embedding that captures the unique timbre, tone, accent, and other characteristics of the speaker, allowing it to replicate the voice in conversation and speech generation. Combined with the above text instructions that define persona, users can easily create new AI characters capable of engaging in natural, interactive conversations. Thanks to the easy customizability, we were able to pre-build millions of diverse voices.

\item {\bf Unified model for various audio tasks}: 
In addition to spoken dialogue, \modelname as a unified model also naturally supports a variety of audio tasks such as ASR and TTS, without requiring task-specific specialization. Besides, it can be easily extended to handle other audio tasks, such as speech translation, through simple finetuning. Trained on large multilingual text and audio data, \modelname supports six languages: English, Chinese, French, German, Japanese, and Korean.

\end{itemize}
\end{addmargin}







\section{Related Work}

\paragraph{Pipeline Systems} 
Early voice assistant systems, such as Siri, Alexa, and Google Assistant, consists of complex multi-stage pipelines. They typically begin with wake-word detection, which constantly listen for trigger phrases (like ``\texttt{Hey Siri}'') to activate the assistant. Next, automatic speech recognition (ASR) converts human speech into text. Natural language understanding (NLU) then analyzes the text to determine user intent and extract relevant information. In response, natural language generation (NLG) composes an appropriate reply, which is then converted back to spoken language through text-to-speech (TTS), allowing the assistant to ``speak'' to the user.
Recent systems integrate LLMs to simplify the pipeline and enable open-ended conversations. For example, HuggingGPT \citep{shen2023hugginggptsolvingaitasks} and AudioGPT \citep{huang2023audiogptunderstandinggeneratingspeech} connect ASR, LLM, and TTS. 
However, this multi-module approach can introduce significant delays, making it unsuitable for low-latency, real-time applications. Additionally, converting audio to text often results in the loss of essential acoustic information such as tone and emotion \citep{faruqui2022revisiting,lin2022duplex}.

\paragraph{End-to-end Models}
End-to-end audio-language models aim to overcome the above limitations \citep{lakhotia2021generativespokenlanguagemodeling,hassid2024textually,rubenstein2023audiopalmlargelanguagemodel}.
Audio modality can be integrated into pretrained LLMs using connector modules that align audio representations with the model's input space. For instance, models in \citep{Qwen-Audio,Qwen2-Audio,li2024baichuanomni,tang2024salmonngenerichearingabilities,shu2023llasm} use Whisper encoder \citep{radford2022whisper} to convert speech signals into embeddings. The speech and text embeddings are then concatenated as inputs to LLMs. 
However, Whisper encoder may introduce significant latency since it requires the full input sequence before processing, making it unsuitable for real-time streaming settings. Additionally, \citet{tan2024ssr} uses speech-text alignments to segment and compress speech features, matching the granularity of text tokens. However, these methods do not support speech generation and are limited to text outputs.

To support speech generation, recent methods encode continuous audio signals into discrete units (i.e., audio tokens) typically derived from self-supervised models \citep{hsu2021hubertselfsupervisedspeechrepresentation,babu2021xls,zhang2023speechtokenizer,liu2024dinosr}. These units are then incorporated into LLM's vocabulary, enabling the modeling of audio as a foreign language using next-token prediction. The audio tokens predicted by the LLM are then decoded back into audio signals as outputs. GSLM \citep{lakhotia2021generativespokenlanguagemodeling}, TWIST \citep{hassid2024textually}, SpeechGPT \citep{zhang2023speechgpt} and VoxtLM \citep{maiti2024voxtlm} use Wav2Vec \citep{schneider2019wav2vec} or HuBERT \citep{hsu2021hubertselfsupervisedspeechrepresentation} to tokenize continuous speech signals into learned discrete tokens. These tokens are incorporated into LLM's vocabulary for standard next-token prediction training. 
AudioLM \citep{borsos2023audiolm} proposed to combine semantic tokens with acoustic tokens from a neural audio codec \citep{zeghidour2021soundstream}, preserving the acoustic information of the input signal and enabling the model to simulate the output of any sound, including non-speech sounds. SpeechTokenizer \citep{zhang2023speechtokenizer} distills from Hubert and unifies the semantic and acoustic tokens by disentangling different aspects of speech information hierarchically. Spectron \citep{nachmani2023spoken} is trained directly on spectrograms without any quantization, which naturally preserves the input's semantic and acoustic characteristics. As Spectron predicts the spectrograms for audio output, it supports both speech and text generation, without altering the LLM vocabulary.

Using multimodal inputs combining speech and text allows the model to learn cross-modal knowledge collaboratively. Recent works \citep{zhang2023speechgpt,nachmani2023spoken,nguyen2024spirit,mitsui2024pslm} further exploit cross-modal information interaction. For example, SpeechGPT and Spectron use a Chain-of-Modality approach, guiding the model to process information textually before converting the generated text to speech. These methods fully leverage the LLM backbone's reasoning abilities acquired during text pretraining. However, this hierarchical structure of speech and text requires the model to output a complete textual response before generating the speech output, leading to increased latency that is not ideal for streaming scenarios. In contrast, Spirit-LM \citep{nguyen2024spirit} and USDM \citep{kim2024integratingparalinguisticsspeechempoweredlarge} address the limitation by using an interleaved format of text and speech tokens, where some of the text tokens in a sequence are replaced with speech tokens. Yet, the text and speech tokens conveying the same semantics usually do not align on a token-by-token basis, making the token replacement noisy. This lack of explicit synchronization between text and speech makes the model training more difficult.
On the other hand, PSLM \citep{mitsui2024pslm} replaces the hierarchical speech-text structure with a parallel approach; however, it depends on an external ASR system to provide text input for the spoken audio.

\paragraph{Full-duplex Models}
End-to-end models still follow mechanical, turn-based conversation dynamics, where one party speaks while the other waits to respond. In contrast, full-duplex models allow simultaneous two-way communication that mirrors natural human interactions. These models can listen and speak at the same time, forming the foundation for autonomous interactions where the system continuously listens and actively participates by initiating speech (e.g., through backchanneling or interrupting) when appropriate.
Moshi \citep{defossez2024moshi} is a full-duplex speech-text model combining several ideas mentioned above, including semantic and acoustic discrete audio tokens, and the interleaved structure of Spirit-LM with the parallel mode of PSLM through an Inner Monologue module. This combination enhances the factual accuracy and linguistic quality of generated speech in streaming mode. However, the Inner Monologue mechanism requires specific configurations to support different tasks, such as spoken dialogue, ASR, and TTS, making it difficult to use one single model to support all applications. For example, in ASR tasks, the audio tokens need to precede the text, whereas in TTS tasks, the text delay is adjusted so that the text appears before the audio tokens. 
Hertz-dev \citep{standardintelligence2024hertzdev} is a pure audio model without leveraging the power of pretrained LLMs. Their ablation studies show pretraining with text data does not bring notable improvement over training only on audio data. In comparison, \modelnameauto offers several unique advantages, including better integration of LLM text capabilities with the new audio capabilities, easy customizability with text instructions and plug-and-play voice embedding, and unified modeling of spoken dialogue, ASR, TTS and other various tasks in a single model.

\section{Voila: Voice-Language Foundation Models}

As shown in Figure~\ref{fig:viola-arch-joint}, \modelname adopts a hierarchical multi-scale transformer-based architecture consisting of a voice-language LLM backbone and an audio transformer. The LLM backbone is used to model the semantic information, while the audio transformer models the audio tokens based on the semantic output of the LLM. The audio tokens generated by the audio transformer are finally decoded back to audio by the \modelname tokenizer (\S\ref{sec:method:tokenizer}). \modelnameete supports end-to-end voice conversations that capture nuanced vocal information. \modelnameauto further extends over \modelnameete and is a full-duplex model, enabling simultaneous listening, reasoning, and speaking in a  two-way communication that emulates natural human interactions. We develop new text-audio alignment methods (\S\ref{sec:method:alignment}) that take full advantages of pretrained LLMs when building the voice-language models. \modelname models allow easy speaker customization via both text instructions and reference voice embeddings (\S\ref{sec:method:voice}).

\subsection{Voice Tokenizer}\label{sec:method:tokenizer}
By transforming continuous audio signals into discrete tokens, LLM can be trained/finetuned with next-token prediction to understand and generate audio. Discrete audio tokens can be classified into two categories, namely semantic tokens \citep{hsu2021hubertselfsupervisedspeechrepresentation,schneider2019wav2vec} and acoustic tokens \citep{zeghidour2021soundstream,défossez2022highfidelityneuralaudio}. HuBERT \citep{hsu2021hubertselfsupervisedspeechrepresentation} obtains semantic tokens by applying K-means clustering to the activation hidden space, which shows effectiveness in capturing high-level linguistic content and supporting language modeling and resynthesis \citep{zhang2023speechgpt,polyak2021speechresynthesisdiscretedisentangled}. 
However, semantic tokens lose acoustic details such as speaker identity, intonation, and emotion, resulting in suboptimal reconstruction. In contrast, acoustic tokens generated by neural codec models \citep{zeghidour2021soundstream,défossez2022highfidelityneuralaudio} with residual vector quantization (RVQ) can effectively restore sound, but they have weak semantic dependency, making it difficult for LLM training/finetuning to converge. 
We rather extend the approach of \citet{zhang2023speechtokenizer} by distilling semantic information into the first level of tokens with RVQ. The first level of RVQ tokens focuses on semantic information and the other three levels learn the acoustic information. We trained our tokenizer with 100K hours of audio data.

\begin{figure}[t]
    \centering
    \includegraphics[width=0.9\textwidth]{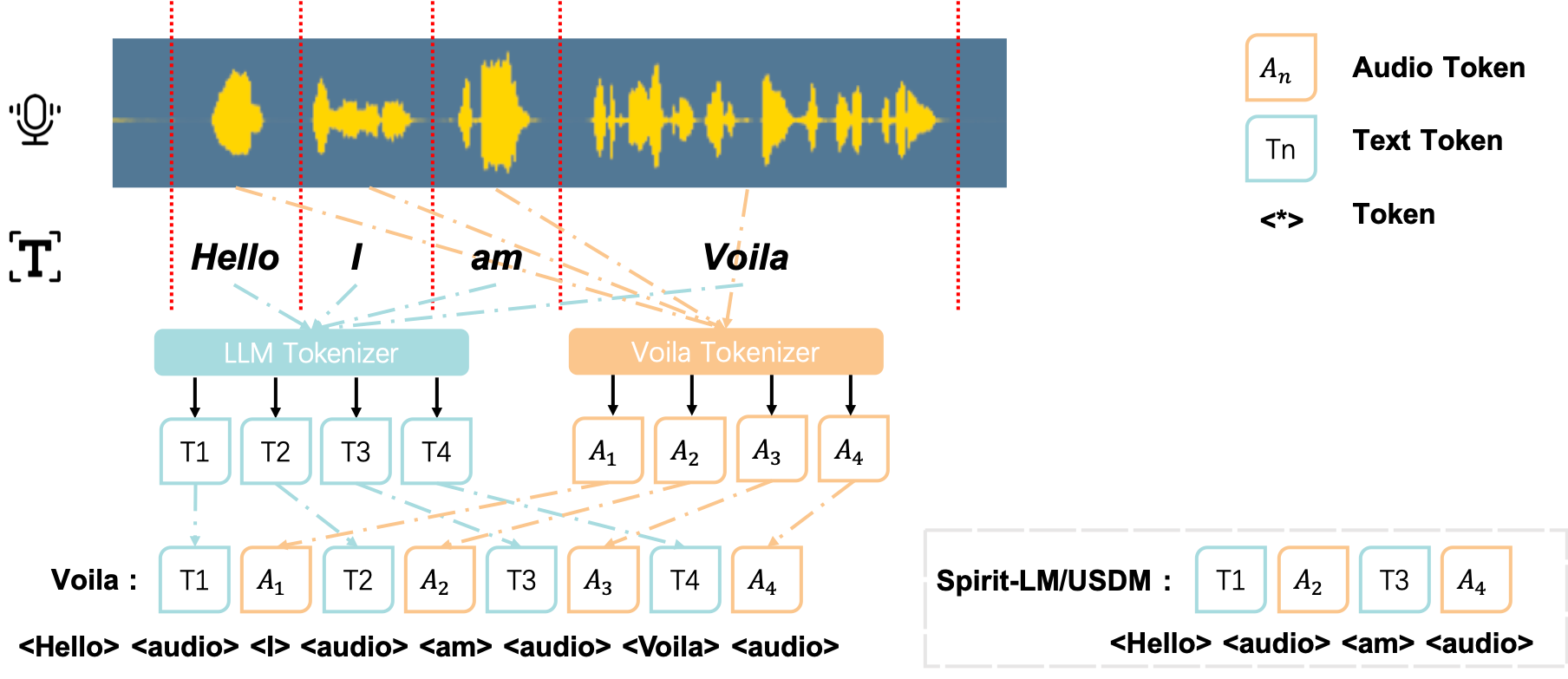}
    \caption{Text and audio interleaved alignment.}
    \label{fig:model_input}
\end{figure}

\subsection{Text and Audio Alignment} \label{sec:method:alignment}

\paragraph{Multi-task alignment. }
We integrate the discrete audio tokens extracted by the above \modelname tokenizer into the vocabulary of the backbone LLM. To align the text and audio modalities, we train the model on tasks including automatic speech recognition (ASR), text-to-speech (TTS), and instruction following. These tasks are unified under a chat-style format, with next-token prediction as the training objective.
For ASR, the input-output sequence is structured as \texttt{'<human> audio input <voila> text output <eos>'}, where the audio input consists of discrete audio tokens and the model generates the corresponding transcript. In TTS, the format is \texttt{'<human> text input <voila> audio output <eos>'}, with the model predicting audio tokens from a given text input. Instruction-following data is expressed in four formats: Text Input $\to$ Text Output (TITO), Text Input $\to$ Audio Output (TIAO), Audio Input $\to$ Text Output (AITO), and Audio Input $\to$ Audio Output (AIAO). For all tasks, we compute the loss only on the response portion, i.e., the tokens between \texttt{'<voila>'} and \texttt{'<eos>'}. When the instruction output involves audio (as in TIAO and AIAO), we adopt an interleaved format of text and audio tokens to improve alignment between modalities, as described below.

\begin{figure}[t]
    \centering
    \includegraphics[width=\textwidth]{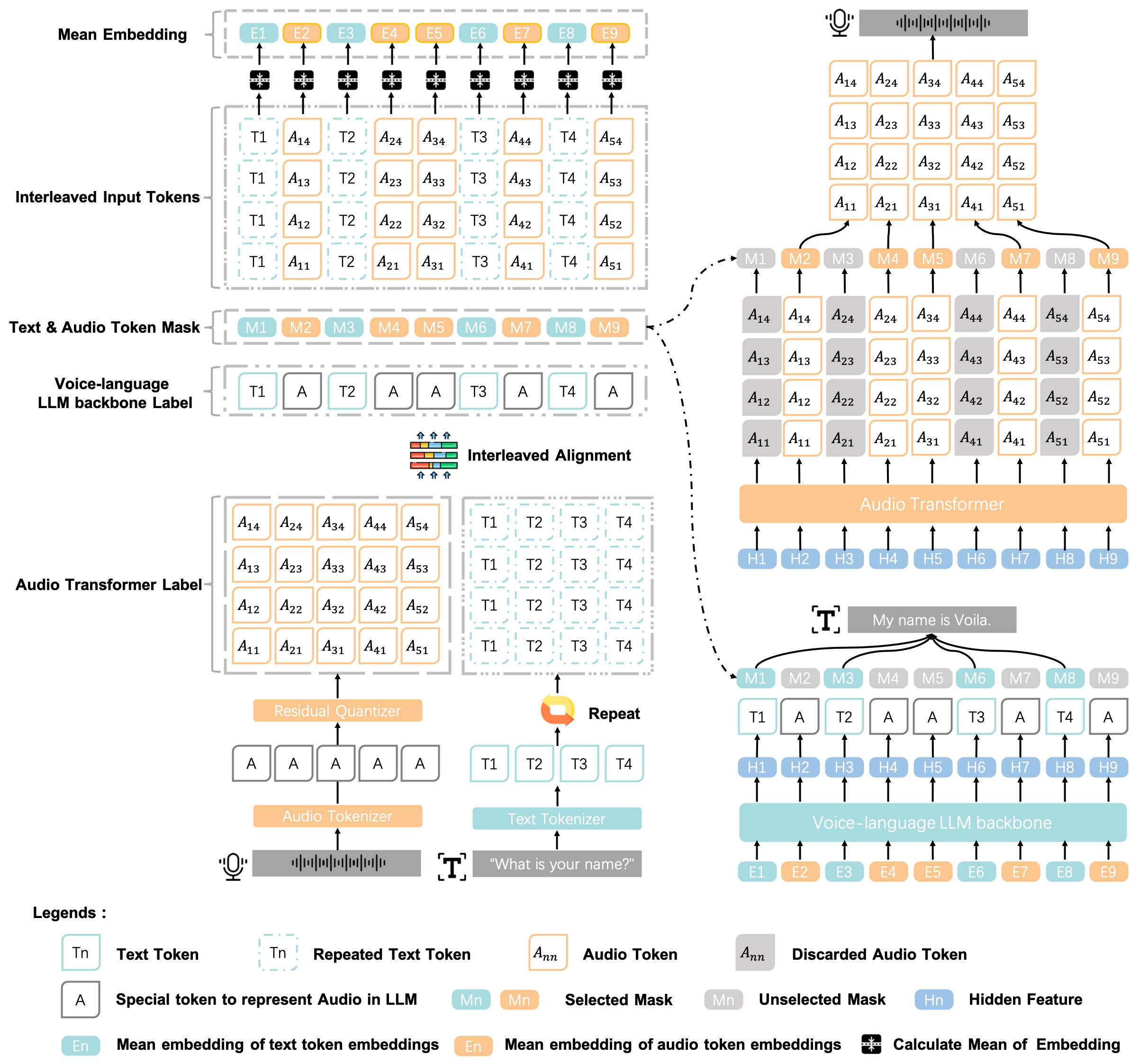}
    \caption{Input embedding and output decoding in \modelname.}
    \label{fig:model_input_and_output}
\end{figure}

\begin{figure}[t]
    \centering
    \includegraphics[width=0.8\linewidth]{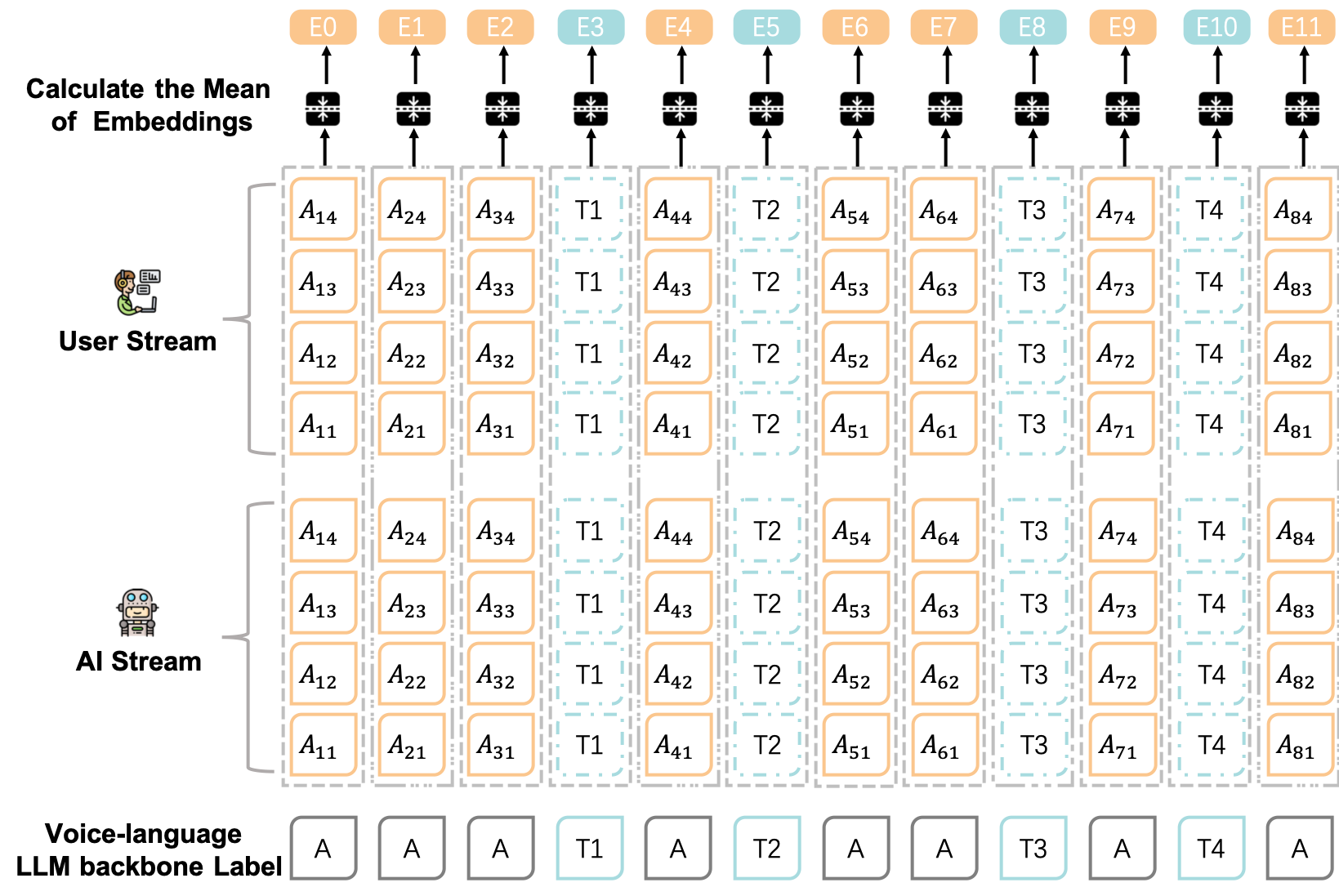}
    \caption{\modelnameauto two-stream inputs, including user’s audio stream and \modelname’s own audio stream.}
    \label{fig:voila_aotonomous}
\end{figure}

\paragraph{Text-audio interleaved alignment. }
To improve alignment between text and audio, \modelname adopts a structured interleaved alignment strategy, where each semantic unit of text is paired with its corresponding audio tokens in an alternating sequence. For example, as illustrated in Figure~\ref{fig:model_input}, given the spoken input \texttt{'Hello I am Voila'}, the input sequence is encoded as \texttt{'<Hello> <audio> <I> <audio> <am> <audio> <Voila> <audio>'}, ensuring that each word is tightly aligned with its corresponding audio segment. This design facilitates fine-grained alignment and enhances the model’s ability to generate expressive and synchronized speech.
This design differs from prior approaches, such as Spirit-LM \citep{nguyen2024spirit} and USDM \citep{kim2024integratingparalinguisticsspeechempoweredlarge}, which also adopt interleaved text-audio formats but do so with looser coupling. These methods alternate between sequences of text and audio tokens without enforcing one-to-one alignment, often requiring the model to infer the correspondence between modalities implicitly (Figure~\ref{fig:model_input}). 
The lack of explicit alignment can hinder training stability and limit the expressiveness of generated speech.


Figure~\ref{fig:model_input_and_output} illustrates the overall input-output architecture. We use audio tokens produced by the four-layer RVQ tokenizer, as described in \S\ref{sec:method:tokenizer}. To maintain dimensional consistency with the audio tokens, each text token is repeated four times (bottom-left panel of the figure). The response segments of instruction-following data---which contain both text and audio---are then arranged into an interleaved pattern as described above. Tokens from this sequence are extracted and converted into embeddings (top-left panel), whose mean is computed and fed into the backbone LLM (bottom-right panel). Finally, the audio transformer takes the output from the backbone LLM as input to predict the corresponding audio tokens (top-right panel).

As shown in Figure~\ref{fig:voila_aotonomous}, \modelnameauto operates as a full-duplex model, processing both the user’s audio stream and \modelname’s own audio stream simultaneously. Each stream is independently tokenized and embedded. Once embeddings from both streams are obtained, they are fused by averaging and then passed into the backbone LLM. Finally, the audio transformer generates the \modelname audio output by modeling the corresponding audio tokens.

\subsection{One Million Pre-built Voices and Customizing New Voices} \label{sec:method:voice}

\modelname allows users to easily customize and plug in new voices for conversations. 
Unlike recent pipeline-based systems that handle voice customization through separate TTS modules \citep{huang2025stepaudiounifiedunderstandinggeneration}, \modelname integrates this functionality directly into a unified, end-to-end and autonomous framework (Figure~\ref{fig:viola-arch-joint}).

To this end, \modelname introduces a learnable special token that captures a speaker’s unique voice characteristics---including timbre, tone, and accent---via a voice embedding. During inference, this token conditions the model to synthesize speech in the desired voice. Specifically, we use Wespeaker \citep{Wang2023} to extract speaker embeddings from all training data with audio outputs.
For training tasks involving audio generation, we prepend three special tokens to the system prompt: one each to indicate the start, reference point, and end of the voice embedding segment. The extracted speaker voice embedding is added to the embedding of the reference token, effectively conditioning the model on speaker identity. To avoid task confusion, we use different token sets for TTS and chat tasks: \texttt{<TTS\_REF\_START><TTS\_REF><TTS\_REF\_END>} for TTS, and \texttt{<CHAT\_REF\_START><CHAT\_REF><CHAT\_REF\_END>} for chat.


At inference time, a voice embedding can be derived from an audio clip of any length---ranging from a few seconds to several hours---using Wespeaker on the fly. This embedding is then passed to \modelname to generate speech in the same voice. Combined with text instructions that define a persona, users can create fully customized AI characters capable of natural and expressive conversations.
Thanks to \modelname’s efficient customizability, we have pre-built over one million diverse voices, empowering users to further personalize voice profiles dynamically at inference.



\section{Experiments}

\begin{figure}[t]
    \centering
    \includegraphics[width=0.9\linewidth]{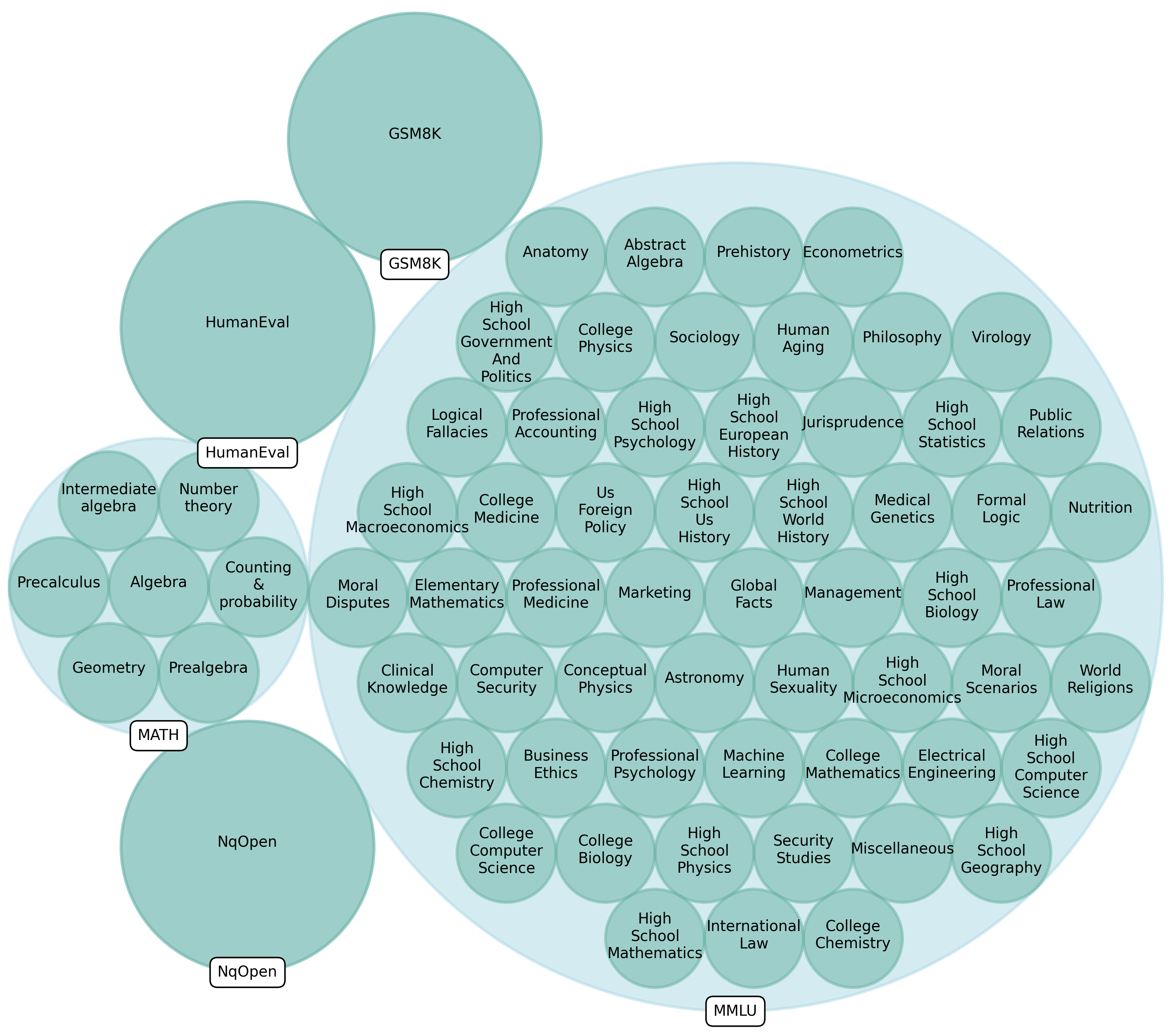}
    \caption{Domain distribution in \modelname Benchmark}
    \label{fig:voila_benchmark}
\end{figure}

\subsection{Voila Benchmark}

Just as recent efforts have established comprehensive benchmarks for evaluating LLMs, a thorough assessment of voice-language models requires evaluation across a wide range of domains. To this end, we introduce the \modelname Benchmark---a new audio-language evaluation suite.
The \modelname Benchmark is constructed by sampling from five widely used LLM evaluation datasets: MMLU \citep{hendryckstest2021}, MATH \citep{hendrycksmath2021}, OpenAI HumanEval \citep{chen2021evaluating}, NQ-Open \citep{nq_open}, and GSM8K \citep{cobbe2021gsm8k}. These samples are then converted into speech using off-the-shelf TTS systems to provide broad domain coverage and realistic audio inputs. Figure~\ref{fig:voila_benchmark} shows the domain distribution of the benchmark.

Specifically, for MMLU, we randomly selected 20 samples from each of its 57 diverse subjects, resulting in 1,140 data points. Similarly, for the MATH dataset, which spans 6 subjects, we selected 20 samples per subject, yielding 120 data points. For OpenAI HumanEval, NQ-Open, and GSM8K, we randomly selected 100 samples from each dataset, treating each as a distinct subject. In total, the \modelname Benchmark comprises 66 subjects and 1,580 samples, all drawn from the test splits of their respective datasets.
Since these evaluation sets originate from text-based sources, some samples---particularly those containing mathematical formulas or code---are not directly suitable for TTS synthesis. To address this, we use GPT-4o (\texttt{gpt-4o-2024-08-06}) to rewrite the text into a TTS-friendly format, which is then converted into speech using Google Cloud's TTS technology.

\begin{figure}[t]
    \centering
    \includegraphics[width=\textwidth]{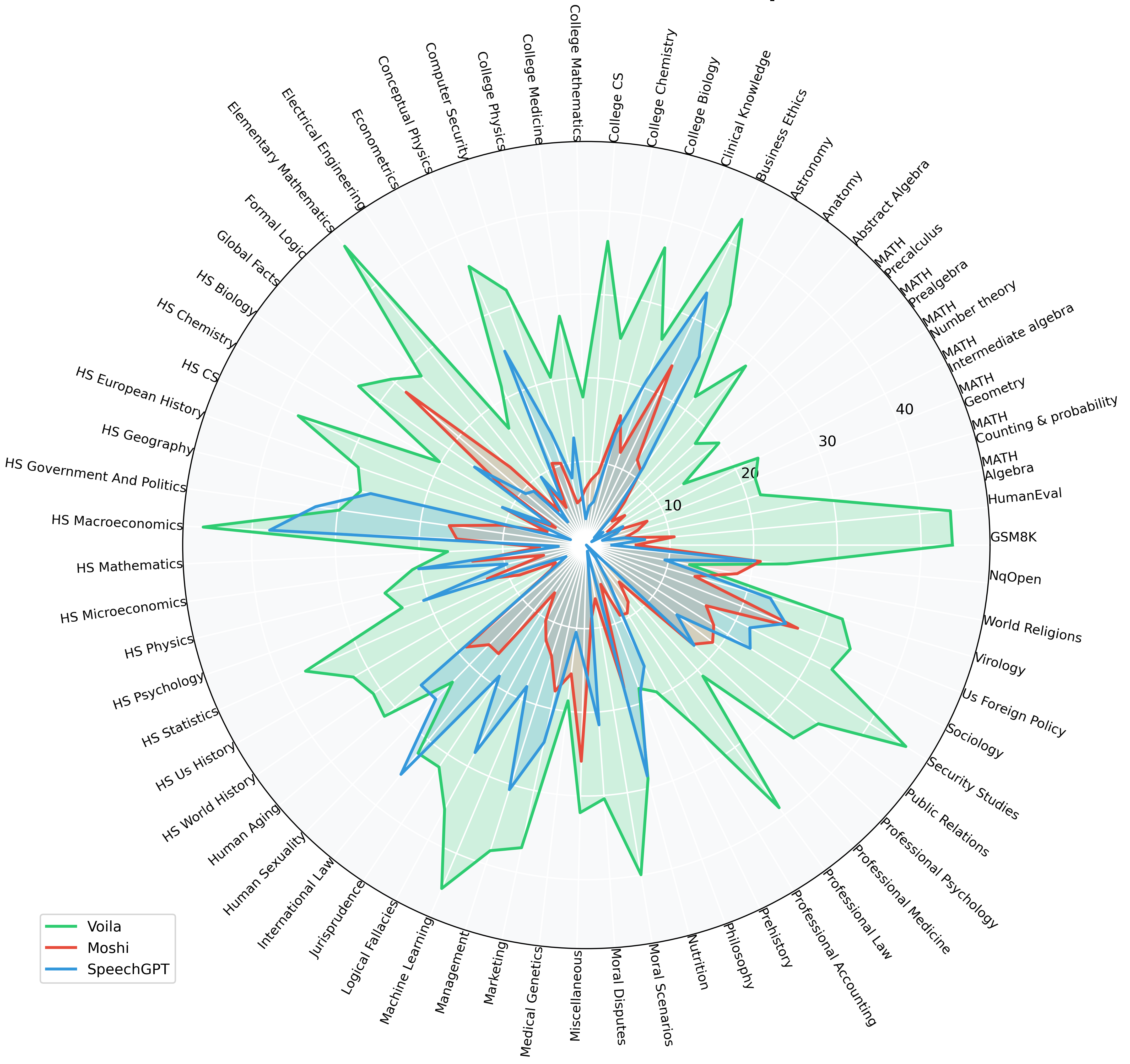}
    \caption{Performance comparison across the diverse domains in \modelname Benchmark.}
    \label{fig:voila_benchmark_radar}
\end{figure}

\begin{table}[t]
    \centering
    \caption{Overall performance on \modelname Benchmark}
    \begin{tabular}{@{}ll@{}} 
        \toprule
        Model & Accuracy \\ 
        \midrule
        SpeechGPT (7B) \citep{zhang2023speechgpt} & 13.29 \\
        Moshi \citep{defossez2024moshi} & 11.45 \\
        \midrule
        Voila & \textbf{30.56} \\
        \bottomrule
    \end{tabular}
    \label{tab:voila_benchmark} 
\end{table}


\subsection{Evaluation on \modelname Benchmark}

To evaluate the correctness of audio outputs from voice-language models, we first transcribe the generated speech using the Whisper system \cite{radford2022whisper}. We then use GPT-4o to assess the transcribed responses. For each test case, GPT-4o is provided with the question and the reference answer as ground truth, and it assigns a score to the model’s output on a scale from 0 to 100 based on its alignment with the reference answer.

We compare our results against two recent open-source audio-language models: SpeechGPT \citep{zhang2023speechgpt} and Moshi \citep{defossez2024moshi}. Table~\ref{tab:voila_benchmark} presents the average scores of all models on the \modelname Benchmark. Detailed performance across the benchmark’s diverse fine-grained domains is shown in Figure~\ref{fig:voila_benchmark_radar} and Table~\ref{tab:voila_benchmark_domain}. \modelname outperforms both SpeechGPT and Moshi, establishing a strong baseline on this benchmark.
Notably, \modelname demonstrates significant improvements in the math and code domains, highlighting \modelname's text-audio alignment takes effective advantage of the reasoning capabilities of the backbone LLM.

\scriptsize{

\begin{longtable}{llll}
    \caption{Detailed results across the diverse domains in \modelname Benchmark.}
    \endfirsthead
    \toprule
        {\bf Domain} & {\bf SpeechGPT} & {\bf Moshi} & {\bf Voila} \\
    \midrule
        MMLU-High School Microeconomics & 20.25 & 13.75 & \textbf{21.00} \\
    \midrule
        MMLU-High School World History & 4.45 & 5.15 & \textbf{31.05} \\
    \midrule
        MMLU-High School Statistics & 4.75 & 8.75 & \textbf{36.80} \\
    \midrule
        MMLU-High School Biology & 16.30 & 12.25 & \textbf{33.20} \\
    \midrule
        MMLU-High School European History & 2.00 & 4.95 & \textbf{28.80} \\
    \midrule
        MMLU-High School US History & 2.80 & 4.25 & \textbf{32.00} \\
    \midrule
        MMLU-High School Psychology & 20.55 & 12.50 & \textbf{23.25} \\
    \midrule
        MMLU-High School Government and Politics & 32.70 & 16.55 & \textbf{29.80} \\
    \midrule
        MMLU-High School Macroeconomics & 37.90 & 15.50 & \textbf{45.85} \\
    \midrule
        MMLU-High School Chemistry & 5.05  & 4.25 & \textbf{20.20}\\
    \midrule
        MMLU-High School Mathematics & 3.35 & 5.60 & \textbf{16.60} \\
    \midrule
        MMLU-High School Computer Science & 11.00 & 11.00 & \textbf{37.75} \\
    \midrule
        MMLU-High School Geography & 26.50 & 10.25 & \textbf{27.75} \\
    \midrule
        MMLU-Medical Genetics & 10.50 & 15.50 & \textbf{18.75} \\
    \midrule
        MMLU-Jurisprudence & 18.80 & 6.85 & \textbf{31.85} \\
    \midrule
        MMLU-Formal Logic & 9.00 & 13.00 & \textbf{28.25} \\
    \midrule
        MMLU-Management & 30.65 & 13.90 & \textbf{38.30} \\
    \midrule
        MMLU-Global Facts & 9.50 & 28.25 & \textbf{30.65} \\
    \midrule
        MMLU-Human Aging & 25.90 & 18.85 & \textbf{31.65} \\
    \midrule
        MMLU-Philosophy & \textbf{18.75} & 5.00 & 18.25 \\
    \midrule
        MMLU-Sociology & 21.90 & 16.05 & \textbf{32.90} \\
    \midrule
        MMLU-Marketing & 24.10 & 17.85 & \textbf{37.00} \\
    \midrule
        MMLU-Human Sexuality & 22.90 & 16.65 & \textbf{25.75} \\
    \midrule
        MMLU-US Foreign Policy & 25.60 & 27.15 & \textbf{33.90} \\
    \midrule
        MMLU-Astronomy & 26.25 & 11.85 & \textbf{33.45} \\
    \midrule
        MMLU-Moral Scenarios & 0.75 & 6.50 & \textbf{39.95} \\
    \midrule
        MMLU-Professional Law & 1.90 & 8.45 & \textbf{38.95} \\
    \midrule
        MMLU-College Mathematics & 3.15 & 6.65 & \textbf{17.70} \\
    \midrule
        MMLU-Machine Learning & 18.35 & 12.40 & \textbf{44.55} \\
    \midrule
        MMLU-Electrical Engineering & 9.75 & 8.85 & \textbf{16.75} \\
    \midrule
        MMLU-Conceptual Physics & 25.15 & 10.60 & \textbf{36.15} \\
    \midrule
        MMLU-Professional Medicine & 0.00 & 5.90 & \textbf{20.95} \\
    \midrule
        MMLU-Professional Psychology & 17.60 & 17.40 & \textbf{33.85} \\
    \midrule
        MMLU-Professional Accounting & 5.25 & 9.50 & \textbf{25.15} \\
    \midrule
        MMLU-College Physics & 8.20 & 5.15 & \textbf{20.50} \\
    \midrule
        MMLU-College Medicine & 12.90 & 5.50 & \textbf{27.55} \\
    \midrule
        MMLU-Computer Security & 14.00 & 10.25 & \textbf{31.95} \\
    \midrule
        MMLU-Security Studies & 23.15 & 18.00 & \textbf{45.15} \\
    \midrule
        MMLU-High School Physics & 9.75 & 5.25 & \textbf{24.75} \\
    \midrule
        MMLU-Business Ethics & 33.40 & 23.75 & \textbf{43.15} \\
    \midrule
        MMLU-Miscellaneous & 14.05 & 25.85 & \textbf{32.00} \\
    \midrule
        MMLU-College Biology & 13.75 & 16.00 & \textbf{36.75} \\
    \midrule
        MMLU-Elementary Mathematics & 3.55 & 5.30 & \textbf{45.95} \\
    \midrule
        MMLU-Prehistory & 16.05 & 9.30 & \textbf{19.45} \\
    \midrule
        MMLU-Logical Fallacies & 28.15 & 10.30 & \textbf{35.85} \\
    \midrule
        MMLU-Abstract Algebra & 5.15 & 6.85 & \textbf{28.65} \\
    \midrule
        MMLU-Econometrics & 6.25 & 5.10 & \textbf{21.50} \\
    \midrule
        MMLU-Clinical Knowledge & 21.00 & 11.80 & \textbf{26.20} \\
    \midrule
        MMLU Anatomy & 11.50 & 11.00 & \textbf{22.00} \\
    \midrule
        MMLU-College Computer Science & 4.65 & 7.80 & \textbf{36.40} \\
    \midrule
        MMLU-Moral Disputes & 21.55 & 8.75 & \textbf{30.40} \\
    \midrule
        MMLU-International Law & \textbf{35.25} & 16.70 & 32.00 \\
    \midrule
        MMLU-Public Relations & 13.65 & 19.05 & \textbf{35.00} \\
    \midrule
        MMLU-Virology & 22.90 & 13.50 & \textbf{31.85} \\
    \midrule
        MMLU-College Chemistry & 5.25 & 8.80 & \textbf{25.05} \\
    \midrule
        MMLU-Nutrition & 28.55 & 17.65 & \textbf{28.90} \\
    \midrule
        MMLU-World Religions & 9.55 & \textbf{18.35} & 12.55 \\
    \midrule
        MATH-Intermediate Algebra & 4.90 & 3.50 & \textbf{23.00} \\
    \midrule
        MATH-Algebra & 4.65 & 4.70 & \textbf{29.00} \\
    \midrule
        MATH-Geometry & 3.90 & 7.80 & \textbf{21.70} \\
    \midrule
        MATH-Number Theory & 0.75 & 3.00 & \textbf{13.75} \\
    \midrule
        MATH-Precalculus & 1.30 & 4.20 & \textbf{17.80} \\
    \midrule
        MATH-Counting Probability & 2.00 & 6.40 & \textbf{21.65} \\
    \midrule
        MATH-Prealgebra & 2.50 & 5.80 & \textbf{20.00} \\
    \midrule
        OpenAI HumanEval & 7.01 & 10.55 & \textbf{43.70} \\
    \midrule
        NQ Open & 20.24 & 20.89 & \textbf{24.07} \\
    \midrule
        GSM8K & 3.05 & 5.96 & \textbf{43.76} \\
    \bottomrule
    \label{tab:voila_benchmark_domain}
\end{longtable}

}

\normalsize

\begin{table}[h]
    \centering
    \caption{Results of ASR.}
    \begin{tabular}{@{}ll@{}} 
        \toprule
        Model & LibriSpeech test-clean (WER $\downarrow$) \\ 
        \midrule
        Whisper large v2 \citep{radford2022whisper} & 2.7 \\ 
        Whisper large v3 \citep{radford2022whisper} & 2.2 \\ 
        FastConformer \citep{rekesh2023fast}(w/ LibriSpeech train split) & 3.6 \\ 
        VoxtLM \citep{maiti2024voxtlmunifieddecoderonlymodels}(w/ LibriSpeech train split) & 2.7 \\ 
        Moshi \citep{defossez2024moshi} & 5.7 \\ 
        \midrule
        Voila (w/o LibriSpeech train split) & \textbf{4.8} \\ 
        Voila (w/ LibriSpeech train split) & \textbf{2.7} \\ 
        \bottomrule
    \end{tabular}
    \label{tab:asr_results} 
\end{table}

\begin{table}[h]
    \centering
    \caption{Results of TTS.}
    \begin{tabular}{@{}ll@{}} 
        \toprule
        Model & LibriSpeech test-clean (WER $\downarrow$) \\ 
        \midrule
        YourTTS \citep{casanova2022yourtts} & 7.7 \\ 
        Vall-E \citep{wang2023neural} & 5.9 \\ 
        Moshi \citep{defossez2024moshi} & 4.7 \\ 
        \midrule
        Voila (w/o LibriSpeech train split) & \textbf{3.2} \\ 
        Voila (w/ LibriSpeech train split) & \textbf{2.8} \\ 
        \bottomrule
    \end{tabular}
    \label{tab:tts_results} 
\end{table}

\subsection{Evaluation on ASR and TTS}

\textbf{Metrics.} 
\modelname supports not only spoken dialogue but also automatic speech recognition (ASR) and text-to-speech (TTS). We evaluate both components independently. For ASR, we measure performance on the LibriSpeech test-clean dataset \citep{panayotov2015librispeech}, using word error rate (WER) as the evaluation metric. For TTS, we follow the protocol from Vall-E \citep{wang2023neural}, transcribing the generated audio with HuBERT-Large \citep{hsu2021hubertselfsupervisedspeechrepresentation}. Following the Vall-E setup, we evaluate audio samples between 4 and 10 seconds in length. We report results under two settings: (1) when LibriSpeech data is excluded during training, and (2) when the LibriSpeech training split is included.

\textbf{Baselines.} 
We compare our ASR results with prior work, including \citet{radford2022whisper, rekesh2023fast, maiti2024voxtlmunifieddecoderonlymodels, defossez2024moshi}. For TTS, we compare with \citet{wang2023neural, defossez2024moshi}.

\textbf{Results.} 
The results in Table~\ref{tab:asr_results} show that \modelname performs competitively with state-of-the-art ASR models such as Whisper \citep{radford2022whisper}, outperforming models like FastConformer \citep{rekesh2023fast}. When compared to recent speech-language models \citep{maiti2024voxtlmunifieddecoderonlymodels, defossez2024moshi}, \modelname achieves a lower WER of 4.8\%, surpassing the 5.7\% reported in \citet{defossez2024moshi}. In the setting where LibriSpeech training data is used, \modelname reaches a WER of 2.7\%, matching the best result reported by \citet{maiti2024voxtlmunifieddecoderonlymodels}. As shown in Table~\ref{tab:tts_results}, \modelname also outperforms in TTS, achieving a WER of 3.2\% (2.8\% with LibriSpeech training data), compared to 4.7\% from \citet{defossez2024moshi}.


\section{Conclusion}

We introduced \modelname, a family of voice-language foundation models that support spoken dialogue, ASR, TTS, and other voice-language tasks in an end-to-end, autonomous manner. Through innovations in voice tokenization, hierarchical modeling, and audio-text alignment, \modelname achieves performance comparable to or exceeding state-of-the-art models. Built on a unique multi-scale Transformer architecture, \modelname is designed to tightly integrate voice and language capabilities, enabling fine-grained processing of both semantic and acoustic signals and fully leveraging the strengths of large language models. A key feature of \modelname is its support for extensive customization, allowing users to create diverse and expressive voice personas that enrich interaction quality. Overall, \modelname represents a step toward autonomous voice AI that acts as a proactive and empathetic partner in human activities. We release the models and code to support further research and development in this space.

\bibliography{references,hu-bib}
\bibliographystyle{abbrvnat}

\end{document}